\documentclass{article}

\PassOptionsToPackage{numbers}{natbib}

\usepackage[preprint]{neurips_2026}
\usepackage{amsmath}
\usepackage{amsthm}
\usepackage{graphicx}
\usepackage{subcaption}
\usepackage{multirow}
\usepackage[ruled,vlined]{algorithm2e}
\newtheorem{theorem}{Theorem}
\newtheorem{lemma}{Lemma}
\theoremstyle{remark}
\newtheorem{remark}{Remark}

\usepackage[utf8]{inputenc} 
\usepackage[T1]{fontenc}    
\usepackage{hyperref}       
\usepackage{url}            
\usepackage{booktabs}       
\usepackage{amsfonts}       
\usepackage{nicefrac}       
\usepackage{microtype}      
\usepackage{xcolor}         
\usepackage{wrapfig}

\title{Jacap: Robust KV Cache Eviction via Jacobian-Based Nonlinear Information Capacity Preservation}

\author{%
  Jiaming Yang$^{1,2}$\hspace{3pt}%
  Chenwei Tang$^{1,2,\ast}$\hspace{3pt}%
  Liangli Zhen$^{3}$\hspace{3pt}%
  Chenyang Zhang$^{1,2}$\hspace{3pt}%
  Jiancheng Lv$^{1,2}$%
  \And
  $^{1}$College of Computer Science, Sichuan University, Chengdu, China \\
  $^{2}$Engineering Research Center of Machine Learning and Industry Intelligence, \\
  Ministry of Education, Chengdu, China \\
  $^{3}$Institute of High Performance Computing, \\
  Agency for Science, Technology and Research (A*STAR), Singapore \\
  $^{\ast}$Corresponding author: \texttt{tangchenwei@scu.edu.cn}
}

\begin{document}

\maketitle

\begin{abstract}
Key-value (KV) cache eviction is essential for scaling long-context inference in Large Language Models. However, existing policies predominantly rely on empirical heuristics, lacking a rigorous characterization of token utility under the inherently nonlinear softmax attention mechanism. In this work, we rethink KV cache eviction through the lens of local information geometry, modeling the attention process as a nonlinear Gaussian communication channel. By performing a first-order Taylor expansion of the attention mapping, we derive the Jacobian Information Capacity, a novel objective that explicitly captures query relevance, softmax sensitivity, and structural diversity. Guided by this theory, we introduce Jacap, a capacity-aware eviction method that utilizes softmax-aware importance weighting and statistical leverage scores for subset selection. Extensive experiments across diverse architectures and benchmarks demonstrate that \textsc{Jacap} delivers superior performance in most scenarios, particularly in high-compression regimes.
\end{abstract}

\section{Introduction}
\label{sec:introduction}

Large language models (LLMs) are increasingly deployed in complex, long-horizon settings such as multi-step reasoning~\citep{plaat2025multi}, tool-augmented agents~\citep{ferrag2025llm}, and interactive decision-making systems~\citep{zhao2025llm}. These applications require processing extremely long contexts, where models must retain and reuse information across thousands or even millions of tokens. Although transformer-based architectures are capable of supporting such tasks in principle, their inference cost grows rapidly with context length, making efficient memory management a critical bottleneck. In particular, the key-value (KV) cache, which stores past attention states for autoregressive decoding, incurs both significant memory overhead and latency as the context grows~\citep{luohe2024kvcache, recasens2025mind}. This has led to extensive research on KV cache compression, spanning techniques such as representation-level quantization and structural pruning~\cite{li2024survey, shi2024keep}. KV cache eviction, which dynamically prunes redundant entries based on importance, has emerged as a particularly promising approach due to its model-agnostic nature and seamless integration into inference pipelines.

KV cache eviction methods operate by assigning an importance score to each cached token and retaining only the top-ranked subset under a given memory budget. Existing methods generally rely on heuristic criteria such as attention scores~\cite{li2024snapkv}, token recency~\cite{xiao2023efficient}, or value norms~\cite{tian2025keepkv} to estimate token importance. While effective in practice, these heuristics lack a principled understanding of which information should be preserved for future queries. As a result, they often treat tokens in isolation and fail to capture the complex redundancy and complementarity across cache entries, leading to inconsistent performance under aggressive compression regimes.

Recent work~\cite{yang2026rethinkingkvcacheeviction} has sought a more rigorous foundation by framing KV cache eviction through the Information Bottleneck (IB) principle~\cite{tishby2000information}. By adopting a linear-gaussian surrogate for attention dynamics, this work defines an information capacity objective that quantifies the mutual information between future queries and attention outputs. This framework provides a unified interpretation of query relevance and output diversity, resulting in practical algorithms based on capacity-aware selection. However, we argue that the linear assumption limits the descriptive power of this framework. Transformer attention is governed by the softmax operator, where normalization and competition effects dictate that a token’s contribution depends not only on its individual representation but also on its relative sensitivity within the total distribution. A linear approximation treats the query-to-output mapping as a fixed transformation, overlooking the saturation and competition dynamics central to information flow in real attention mechanisms.

In this work, we revisit KV cache eviction from a nonlinear information-theoretic perspective. We model the retained attention mechanism as a nonlinear information channel and derive a local approximation of its information capacity using the Jacobian of the attention mapping around a representative query distribution. This formulation explicitly incorporates query-key alignment, softmax competition, and value-space diversity into a unified objective, with linear capacity emerging as a special case under a linear channel approximation.

Building on this insight, we propose \textsc{Jacap}, a practical eviction method that approximates nonlinear capacity using a softmax-sensitive weighting scheme combined with capacity-based selection. Extensive experiments across diverse architectures and benchmarks demonstrate that \textsc{Jacap} delivers superior performance in most scenarios, especially under high compression ratios. 

Our contributions are summarized as follows:
\begin{enumerate}
    \item We propose a nonlinear information-theoretic framework for KV cache eviction, modeling retained attention as a nonlinear channel and deriving a Jacobian-based local capacity objective.
    \item We develop \textsc{Jacap}, a practical eviction algorithm that incorporates softmax sensitivity into capacity-aware token selection.
    \item We empirically demonstrate consistent improvements over prior methods on LongBench, NIAH and AIME25 benchmarks, especially under high compression ratio.
\end{enumerate}

\section{Related Works}
\label{sec:related_works}

\subsection{Efficient LLM Inference}

The rapidly growing computational cost of LLM inference has motivated optimizations across multiple levels of the inference stack. At the representation level, quantization methods such as TurboQuant~\cite{zandieh2025turboquant}, PolarQuant~\cite{han2025polarquant}, and KVQuant~\cite{hooper2024kvquant} reduce memory footprint by storing KV Cache in low-precision formats, though they may introduce non-negligible degradation under extreme long-context settings. At the sequence level, token reduction techniques, including token merging methods such as D2O~\cite{wan2024d2o} and CAM~\cite{zhang2024cam}, aim to shorten the effective context length by consolidating redundant tokens. At the system level, memory management optimizations such as PagedAttention~\cite{kwon2023efficient} and FlashDecoding++~\cite{hong2023flashdecodingpp} improve hardware utilization by reducing KV cache fragmentation and increasing token throughput.

Orthogonal to these approaches, structural methods directly compress the KV cache by selectively retaining tokens during decoding. Our method, \textsc{Jacap}, belongs to this category and is complementary to quantization and system-level optimizations.

\subsection{KV Cache Eviction}

\paragraph{Heuristic-based Eviction.}
KV cache eviction methods aim to alleviate the memory bottleneck by selectively pruning cached tokens during autoregressive decoding. Early approaches primarily rely on heuristic importance metrics derived from observed token behavior. Attention-based methods, such as \textsc{H2O}~\cite{zhang2023h2o} and \textsc{SnapKV}~\cite{li2024snapkv}, prioritize tokens that receive consistently high attention weights across decoding steps. Other approaches emphasize structural diversity: \textsc{KeyDiff}~\cite{park2025keydiff} promotes geometric dissimilarity among keys to reduce redundancy, while \textsc{Knorm}~\cite{devoto2024knorm} uses $l_2$-norms as lightweight proxies for token salience. Recent work has explored estimating token importance beyond observed attention patterns. Expected Attention~\cite{devoto2025expectedattention} approximates the expected attention of future queries under a distributional assumption, enabling importance scoring without access to explicit attention matrices.  

Despite their empirical effectiveness, these methods typically score tokens independently and therefore fail to capture higher-order interactions such as redundancy and complementarity among retained tokens.

\paragraph{Information-Theoretic Perspectives.}
Recent work has introduced information-theoretic formulations for KV-cache eviction based on the Information Bottleneck (IB) principle~\cite{tishby2000information,conklin2026learningforgettingllmtraining}. In particular, \textsc{CapKV}~\cite{yang2026rethinkingkvcacheeviction} formulates token selection as maximizing the mutual information between future queries and attention outputs. Under a linear-Gaussian approximation, the objective reduces to a tractable log-determinant form.

However, this formulation relies on a linear approximation of the attention mechanism, effectively treating the query-to-output mapping as a fixed channel. This simplification neglects the nonlinear normalization and competition effects induced by softmax attention, which fundamentally govern how information is distributed across tokens. In contrast, our work models attention as a nonlinear information channel and derives a Jacobian-based local capacity objective, enabling a more faithful characterization of information flow in the KV cache.

\section{Methodology}
\label{sec:methodology}

In this section, we first revisit the core principles behind KV cache eviction. We then introduce a local information-capacity framework that accounts for the nonlinear sensitivity of softmax attention. Based on this objective, we propose \textsc{Jacap} (\textbf{Ja}cobian-\textbf{Cap}acity Aware Eviction), a practical KV cache eviction policy.

\subsection{Preliminary}

During autoregressive decoding, each step $t$ attends to all previously generated tokens. Given the current hidden state $h_t$, the model computes
$$
q_t=W_Qh_t,\qquad k_i=W_Kh_i,\qquad v_i=W_Vh_i,\qquad i\le t.
$$
Let $d_k$ be the query/key dimension and $d_o$ be the output dimension. The attention output is
$$
o_t=\mathrm{Attn}(q_t)=W_O\sum_{i\le t}\alpha_{t,i}v_i,
\qquad
\alpha_{t,i}=\frac{\exp(q_t^\top k_i/\sqrt{d_k})}
{\sum_{j\le t}\exp(q_t^\top k_j/\sqrt{d_k})}.
$$
Modern inference systems maintain a KV cache $\{(k_i,v_i)\}_{i=1}^{t}$ to avoid recomputing past keys and values. However, the memory cost of the cache grows linearly with the sequence length, making KV cache eviction necessary for long-context inference.

From a functional perspective, full attention defines a nonlinear mapping from query to output:
$$
f(q)=U\alpha(q),\qquad \alpha(q)=\mathrm{softmax}(Kq/\sqrt{d_k}),
$$
where $u_i=W_Ov_i$, $U=[u_1,\ldots,u_t]\in\mathbb R^{d_o\times t}$, and $K=[k_1^\top;\ldots;k_t^\top]\in\mathbb R^{t\times d_k}$. Given a memory budget $B<t$, eviction selects a subset $\mathcal C\subseteq\mathcal H_t:=\{1,\ldots,t\}$ with $|\mathcal C|\le B$, resulting in a retained cache
$$
\mathcal Z_{\mathcal C}=\{(k_i,v_i)\}_{i\in\mathcal C}.
$$
For the subset $\mathcal C$, define
$$
K_{\mathcal C}=[k_i^\top]_{i\in\mathcal C}\in\mathbb R^{|\mathcal C|\times d_k},
\qquad
U_{\mathcal C}=[u_i]_{i\in\mathcal C}\in\mathbb R^{d_o\times |\mathcal C|}.
$$
The compressed cache induces the restricted mapping
$$
f_{\mathcal C}(q)=U_{\mathcal C}\alpha_{\mathcal C}(q),
\qquad
\alpha_i^{\mathcal C}(q)=
\frac{\exp(q^\top k_i/\sqrt{d_k})}
{\sum_{j\in\mathcal C}\exp(q^\top k_j/\sqrt{d_k})}.
$$
Thus, KV cache eviction can be formulated as a constrained function approximation problem:
$$
\min_{\mathcal C:|\mathcal C|\le B}
\mathbb E_{q\sim P(q)}\big[\|f(q)-f_{\mathcal C}(q)\|_2^2\big],
$$
where $P(q)$ denotes the distribution of future queries.
\begin{figure}[tbp]
  \centering
  \includegraphics[width=\linewidth]{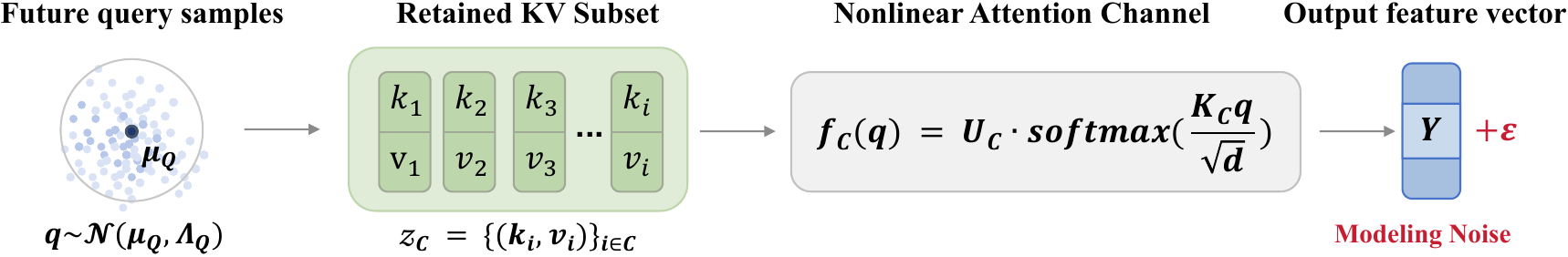}
  \caption{
  \textbf{Nonlinear retained attention channel.}
A retained KV subset $\mathcal Z_{\mathcal C}$ defines a nonlinear softmax attention map from future queries $q\sim\mathcal N(\mu_Q,\Lambda_Q)$ to output features, 
$Y_{\mathcal C}=f_{\mathcal C}(q)+\epsilon$. 
  }
\label{fig:nonlinear_model}
\end{figure}
\subsection{An Information Bottleneck Perspective on KV Cache Eviction}

The Information Bottleneck (IB) principle provides a formal view of the trade-off between compression and predictive information~\citep{tishby2015ib, chechik2003information}. In KV cache eviction, however, the representations are produced by a frozen model, and compression is imposed through a hard cache budget rather than a learned encoder. Therefore, instead of optimizing the full IB Lagrangian, we focus on the predictive information preserved by the retained cache.

Given the retained cache $\mathcal Z_{\mathcal C}$, we model the future query $q$ as a random variable and define the stochastic channel output as
$$
Y_{\mathcal C}=f_{\mathcal C}(q)+\epsilon,
\qquad
\epsilon\sim\mathcal N(0,\Sigma_{\mathrm{noise}}),
\qquad q\perp\epsilon.
$$
The information capacity of the retained cache is then
\begin{equation}
\label{eq:ic}
\mathcal L_{\mathcal C}=I(q;Y_{\mathcal C}\mid\mathcal Z_{\mathcal C}).
\end{equation}

Prior work~\citep{yang2026rethinkingkvcacheeviction} instantiates this objective with a linear-gaussian surrogate,
$$
Y_{\mathcal C}^{\mathrm{Lin}}=U_{\mathcal C}K_{\mathcal C}q+\epsilon,
\qquad
q\sim\mathcal N(\mu_Q,\Lambda_Q).
$$
Under this model, the capacity admits the closed form
\begin{equation}
\label{eq:linear_ic}
\mathcal L_{\mathcal C}^{\mathrm{Lin}}
=
\frac{1}{2}\log\det\left(
I+
\Sigma_{\mathrm{noise}}^{-1}
U_{\mathcal C}K_{\mathcal C}\Lambda_QK_{\mathcal C}^{\top}U_{\mathcal C}^{\top}
\right).
\end{equation}

\subsection{Jacobian Information Capacity for KV Cache Selection}

The linear--Gaussian surrogate in the previous section provides a tractable capacity objective for KV cache selection. 
However, the actual attention computation induced by a retained cache is nonlinear, since the softmax operator normalizes token logits and introduces competition among cached entries. 
Therefore, instead of directly treating the retained cache as a fixed linear channel, we first formulate it as a nonlinear query-to-output mapping.

Given a retained KV subset 
$\mathcal Z_{\mathcal C}=\{(k_i,v_i)\}_{i\in\mathcal C}$, 
we denote by $K_{\mathcal C}$ the matrix of retained keys and by $U_{\mathcal C}$ the corresponding value-induced output directions. 
The retained attention map is defined as
\begin{equation}
\label{eq:nonlinear_retained_map}
f_{\mathcal C}(q)
=
U_{\mathcal C}
\operatorname{softmax}
\left(
\frac{K_{\mathcal C}q}{\sqrt{d_k}}
\right).
\end{equation}
We then model the output representation produced by the retained cache as a noisy nonlinear channel:
\begin{equation}
\label{eq:nonlinear_channel}
Y_{\mathcal C}
=
f_{\mathcal C}(q)+\epsilon,
\qquad
\epsilon\sim\mathcal N(0,\Sigma_{\mathrm{noise}}).
\end{equation}
As illustrated in Fig.~\ref{fig:nonlinear_model}, this formulation preserves the nonlinear softmax normalization that is absent from the global linear surrogate. 
The retained cache no longer defines independent linear channels, but also coupled nonlinear attention channel whose output depends on both query--key accessibility and token-token competition.

To obtain a tractable local capacity measure, we assume that future queries concentrate around a distribution center $\mu_Q$:
$$
q=\mu_Q+\delta q,
\qquad
\delta q\sim\mathcal N(0,\Lambda_Q).
$$
Under this local query model, the nonlinear map in Eq.~\eqref{eq:nonlinear_retained_map} can be approximated by its first-order Taylor expansion around $\mu_Q$.

\begin{figure}[tbp]
  \centering
  \includegraphics[width=\linewidth]{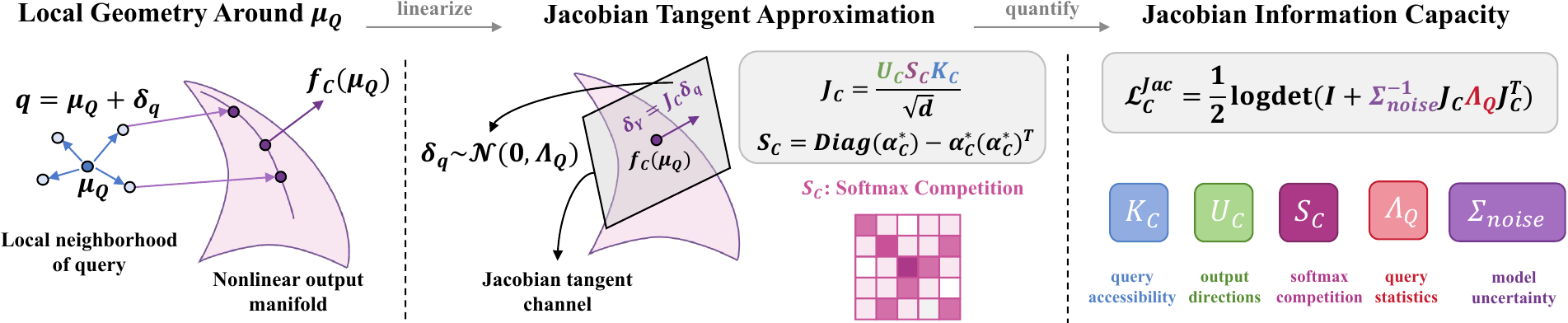}
  \caption{\textbf{Jacobian Information Capacity.}
Around the center $\mu_Q$ of the future query distribution, the nonlinear retained attention map is locally approximated by its Jacobian tangent channel. 
The Jacobian 
$J_{\mathcal C}=U_{\mathcal C}S_{\mathcal C}K_{\mathcal C}/\sqrt{d_k}$ 
decomposes the local channel into key accessibility $K_{\mathcal C}$, output directions $U_{\mathcal C}$, and the softmax competition matrix $S_{\mathcal C}$. 
This yields a local information-capacity objective 
$\mathcal L_{\mathcal C}^{\mathrm{Jac}}$, which quantifies how much query-dependent information is preserved by the retained cache under query statistics $\Lambda_Q$ and modeling uncertainty $\Sigma_{\mathrm{noise}}$.}
  \label{fig:jacobian_capacity}
\end{figure}

\begin{lemma}[Local Jacobian Linearization]
\label{lemma:local_linear}
For a retained cache $\mathcal Z_{\mathcal C}$, the nonlinear attention map $f_{\mathcal C}(q)$ admits the first-order expansion
\begin{equation}
\label{eq:taylor_expand}
f_{\mathcal C}(q)=f_{\mathcal C}(\mu_Q)+J_{\mathcal C}\delta q+\mathcal O(\|\delta q\|_2^2),
\end{equation}
where $J_{\mathcal C}\in\mathbb R^{d_o\times d_k}$ is
\begin{equation}
\label{eq:jacobian_matrix}
J_{\mathcal C}=\frac{1}{\sqrt{d_k}}U_{\mathcal C}S_{\mathcal C}K_{\mathcal C},
\quad
S_{\mathcal C}=\mathrm{Diag}(\alpha_{\mathcal C}^{\star})
-\alpha_{\mathcal C}^{\star}(\alpha_{\mathcal C}^{\star})^\top,
\quad
\alpha_{\mathcal C}^{\star}=\mathrm{softmax}(K_{\mathcal C}\mu_Q/\sqrt{d_k}).
\end{equation}

\end{lemma}

Treating $J_{\mathcal C}$ as the effective local channel gain yields the Jacobian-capacity objective.

\begin{theorem}[Jacobian Information Capacity]
\label{theorem:jacobian_logdet}
Under the local Gaussian perturbation model above and the first-order approximation in Lemma~\ref{lemma:local_linear}, the retained cache capacity is approximated by

\begin{equation}
    \label{eq:jacobian_logdet}
    \begin{split}
        \mathcal L_{\mathcal C}^{\mathrm{Jac}}
        & \approx
        \frac{1}{2}\log\det\left(
        I+\Sigma_{\mathrm{noise}}^{-1}
        J_{\mathcal C}\Lambda_QJ_{\mathcal C}^{\top}
        \right) \\
        & =
        \frac{1}{2}\log\det\left(
        I+
        \frac{1}{d_k}\Sigma_{\mathrm{noise}}^{-1}
        U_{\mathcal C}S_{\mathcal C}K_{\mathcal C}\Lambda_QK_{\mathcal C}^{\top}S_{\mathcal C}U_{\mathcal C}^{\top}
        \right).
    \end{split}
\end{equation}
\end{theorem}

The key difference from the linear capacity in Eq.~\eqref{eq:linear_ic} lies in the softmax Jacobian $S_{\mathcal C}$. 
Its diagonal entries measure the local sensitivity of each token's attention probability to its own logit, while its off-diagonal entries encode the normalization-induced competition between different tokens. 
Therefore, $\mathcal L_{\mathcal C}^{\mathrm{Jac}}$ does not merely reward keys that are aligned with future queries, it rewards retained subsets whose output directions remain locally distinguishable under the nonlinear softmax response. 
As illustrated in Fig.~\ref{fig:jacobian_capacity}, the resulting capacity decomposes into five interpretable components: key accessibility, output directions, softmax competition, query statistics, and model uncertainty.

\begin{remark}
\textit{Under standard smoothness and non-degenerate noise assumptions, the Jacobian approximation matches the first-order local channel induced by nonlinear attention. The remaining approximation error is governed by higher-order Taylor residuals. We provide the detailed derivation and analysis of Theorem~\ref{theorem:jacobian_logdet} in Appendix~\ref{sec:ana_jic}.}

\end{remark}

\subsection{\textsc{Jacap}: A Jacobian-Capacity Aware Eviction Policy}

\begin{algorithm}[htbp]
\caption{\textsc{Jacap}: Jacobian-Capacity-Aware KV Eviction}
\label{alg:jacap}
\SetKwInOut{KwIn}{Require}
\SetKwInOut{KwOut}{Output}

\KwIn{KV cache $\{(k_i,v_i)\}_{i=1}^N$, cache budget $B$, 
query statistics $(\mu_Q, \Lambda_Q)$, temperature $\tau$}
\KwOut{Updated retained KV subset $\mathcal{C}$ with $|\mathcal{C}|=B$}

$\alpha_i^\star \gets \mathrm{softmax}\big(K \mu_Q/\tau\sqrt{d_k}\big)_i$ \tcp*{Compute attention weights at query center}

$s_i \gets \alpha_i^{\star 2}(1-\alpha_i^\star)^2 \cdot (k_i^\top \Lambda_Q k_i)$ \tcp*{softmax-sensitive weight}

$A \gets I + \sum_{i=1}^N s_i \, u_i u_i^\top$ \tcp*{jacobian capacity matrix}

$\text{score}_i \gets s_i \, u_i^\top A^{-1} u_i$ \tcp*{marginal Jacobian-capacity contribution}

$\mathcal{C} \gets \text{TOP-B indices according to score}_i$

\Return $\mathcal{C}$
\end{algorithm}

While Theorem~\ref{theorem:jacobian_logdet} provides a high-fidelity objective for measuring cache utility, directly optimizing this subset-dependent log-determinant objective during online inference is computationally prohibitive. 
Our goal is therefore not to optimize the full Jacobian capacity exactly, but to preserve its essential structure in a lightweight eviction score. 
Following the decomposition in Fig.~\ref{fig:jacobian_capacity}, we retain three ingredients that are most relevant for token selection: local query relevance, softmax sensitivity, and output-space diversity.

The first difficulty comes from the subset-dependent local attention distribution $\alpha_{\mathcal C}^{\star}$. Since the retained subset $\mathcal C$ is unknown before eviction, we compute the local attention distribution over the pre-eviction candidate pool $\mathcal H_t$:
$$
\bar\alpha_i=
\frac{\exp(k_i^\top\mu_Q/\tau\sqrt{d_k})}
{\sum_{j\in\mathcal H_t}\exp(k_j^\top\mu_Q/\tau\sqrt{d_k})},
\qquad i\in\mathcal H_t.
$$
Here $\tau$ denotes the temperature. We use $\bar\alpha_i$ as a subset-independent approximation to the local attention weight of token $i$.

The second difficulty comes from the dense softmax competition matrix $S_{\mathcal C}$, which couples all tokens in the retained subset. For efficient scoring, we retain its diagonal sensitivity and approximate
$$
S_{\mathcal C}\approx \mathrm{Diag}(\rho_i),
\qquad
\rho_i=\bar\alpha_i(1-\bar\alpha_i).
$$
This approximation keeps the local sensitivity induced by softmax saturation while avoiding pairwise token coupling. Under this approximation, the local Jacobian is reduced to a weighted key-output interaction.

We further approximate the query-response covariance $K\Lambda_QK^\top$ by its diagonal entries
$$
\kappa_i=k_i^\top\Lambda_Qk_i,
$$
which measure the variance of the $i$-th key response under future query perturbations. Combining the diagonal softmax sensitivity and the diagonal query covariance yields the nonlinear importance weight
\begin{equation}
\label{eq:jacap_weight}
w_i
=\frac{1}{d_k}\rho_i^2\kappa_i
=\frac{1}{d_k}\bar\alpha_i^2(1-\bar\alpha_i)^2\kappa_i.
\end{equation}
The weight $w_i$ accounts for three local factors: query relevance through $\bar\alpha_i$, softmax sensitivity through $\bar\alpha_i(1-\bar\alpha_i)$, and query-space variability through $\kappa_i$. 
This form differs from pure attention-based scoring: tokens with extremely large attention may become saturated and thus contribute limited local sensitivity, while tokens with moderate but responsive attention can carry larger Jacobian capacity.

With these approximations, the Jacobian capacity reduces to the log-determinant of a weighted output-space capacity matrix. For one-shot selection, we construct this matrix over the full candidate pool:
\begin{equation}
\label{eq:jacap_matrix}
A_{\mathrm{Jac}}
=I+
\sum_{j\in\mathcal H_t}w_j u_j u_j^\top.
\end{equation}
This matrix summarizes the output directions already represented by the candidate cache. Applying the matrix determinant lemma gives the final importance score for each token:
\begin{equation}
\label{eq:jacap_score}
r_i^{\mathrm{Jac}}
=w_i\,u_i^\top A_{\mathrm{Jac}}^{-1}u_i.
\end{equation}

In practice, $\mu_Q$ and $\Lambda_Q$ are estimated from historical queries observed during decoding. When applying the output projection is undesirable, we use $u_i=v_i$ as a lightweight proxy for $W_Ov_i$, following prior capacity-based eviction methods. The full procedure is summarized in Algorithm~\ref{alg:jacap}. We provide the detailed derivation of \textsc{Jacap} together with its computational complexity analysis in Appendix~\ref{app:ana_jacap}. We further conduct empirical runtime evaluations, which show that \textsc{Jacap} achieves comparable efficiency to existing mainstream methods in practice. Detailed runtime comparisons are reported in Appendix~\ref{app:runtime_ana}.

\section{Experiment}
\label{sec:experiment}

\begin{figure}[htbp]
  \centering
  \includegraphics[width=\linewidth]{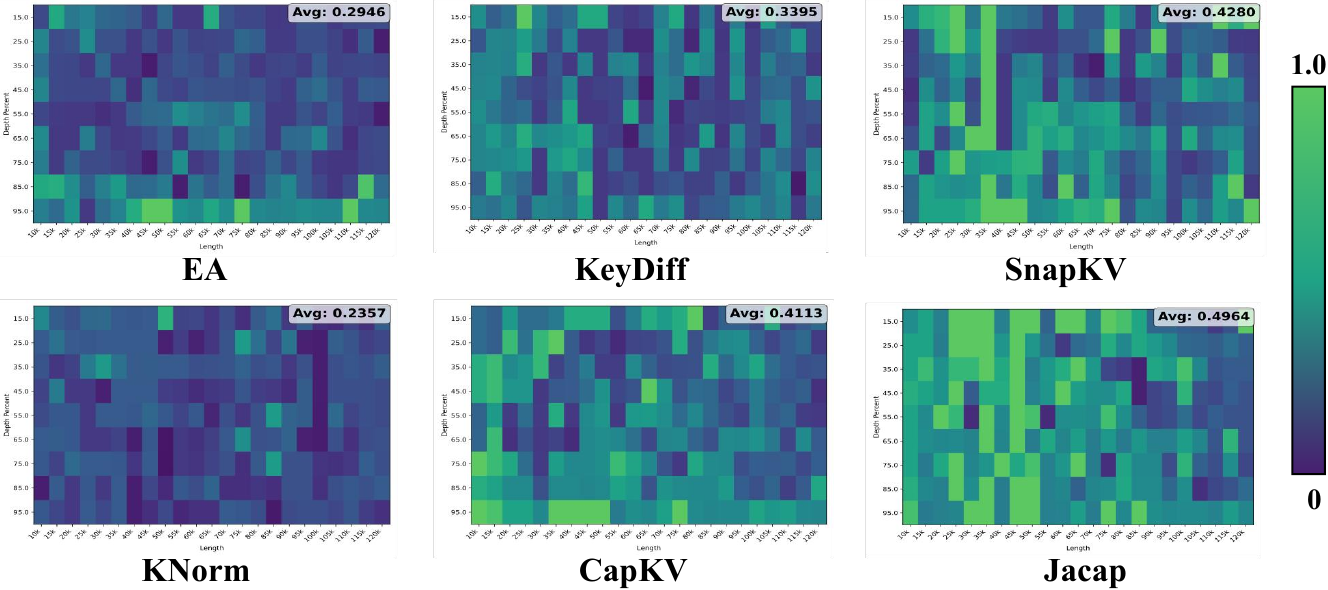}
  \caption{
  Performance of \textsc{Jacap} and baseline KV-cache eviction methods on the Needle-in-Haystack (NIAH) benchmark at compression ratio 0.75 using Qwen3-8B.
  }
\label{fig:niah_75}
\end{figure}

We evaluate our proposed \textsc{Jacap} eviction strategy against representative baselines, including \textsc{CapKV}~\cite{yang2026rethinkingkvcacheeviction}, \textsc{SnapKV}~\cite{li2024snapkv}, \textsc{KeyDiff}~\cite{park2025keydiff}, Expected Attention (\textsc{EA})~\cite{devoto2025expectedattention}, and \textsc{KNorm}~\cite{devoto2024knorm}. The experiments are designed to assess the effectiveness of KV-cache compression in preserving predictive information, handling extremely long contexts, and supporting online decoding under dynamic eviction. 


\subsection{Experiments on LongBench}

\begin{table}[htbp]
  \caption{Performance of \textsc{Jacap} and baseline KV-cache eviction methods on LongBench across different compression ratios (C.R.). For each metric, \textbf{bold} indicates the best-performing method, while \underline{underlined} indicates the second-best.}
  \begin{center}
    \begin{scriptsize}
    \renewcommand{\arraystretch}{0.9}
    \setlength{\tabcolsep}{1.3pt}
      \begin{sc}
        \begin{tabular}{llccccccccccccccccc}
          \toprule
        \toprule
           &  & \multicolumn{3}{c}{SD-QA} & \multicolumn{3}{c}{MD-QA} & \multicolumn{3}{c}{Summ.} & \multicolumn{3}{c}{FSL} & \multicolumn{2}{c}{Synth.} & \multicolumn{2}{c}{Code} & \multicolumn{1}{c}{} \\
          \cmidrule(lr){3-5} \cmidrule(lr){6-8} \cmidrule(lr){9-11} \cmidrule(lr){12-14} \cmidrule(lr){15-16} \cmidrule(lr){17-18} \cmidrule(lr){19-19}
           & C.R. & NQA & Qser & MF & HpQA & 2WQA & Mque & GR & QMS & MN & TREC & TQA & SSum & PC & PR & Lcc & RBP & Avg. \\
          \midrule
          \multicolumn{2}{l}{\textbf{Qwen3-8B}} & 28.89 & 43.57 & 54.76 & 62.86 & 49.20 & 35.69 & 33.67 & 24.60 & 24.99 & 41.00 & 90.21 & 39.95 & 10.50 & 91.02 & 66.83 & 61.95 & 47.48 \\
          \midrule
          \multirow{4}{*}{CapKV} & 0.25 & 28.30 & \textbf{44.13} & \textbf{54.26} & \textbf{62.95} & \textbf{48.47} & \textbf{34.93} & 33.55 & \underline{24.21} & \textbf{25.05} & \textbf{72.00} & \underline{88.47} & \underline{41.00} & 10.50 & \textbf{95.54} & \underline{65.60} & \textbf{63.23} & \textbf{49.51} \\
           & 0.5 & \textbf{29.43} & \underline{41.84} & \textbf{51.11} & 61.15 & 44.40 & \textbf{34.08} & 33.09 & \underline{24.25} & 24.59 & \underline{66.50} & \underline{87.49} & 39.64 & \textbf{11.00} & \textbf{98.42} & 60.06 & \textbf{63.32} & \underline{48.15} \\
           & 0.75 & \textbf{28.62} & \underline{37.69} & \textbf{44.16} & 54.33 & 38.44 & \textbf{30.54} & 30.99 & \textbf{23.58} & 23.12 & 60.50 & \textbf{85.67} & 38.29 & \underline{9.50} & \textbf{96.96} & 51.92 & \textbf{63.81} & \underline{44.88} \\
           & 0.9 & \textbf{24.60} & 27.49 & \textbf{36.12} & \underline{47.69} & 31.49 & \underline{25.12} & 27.11 & \underline{21.61} & 20.27 & 36.00 & \underline{85.97} & 35.00 & 6.50 & \underline{57.38} & 39.65 & \textbf{63.06} & \underline{36.57} \\
          \midrule
          \multirow{4}{*}{EA} & 0.25 & \textbf{28.69} & 42.84 & 53.02 & 62.23 & \underline{48.37} & \underline{33.99} & \underline{33.61} & \textbf{24.22} & \underline{24.94} & 57.00 & 88.13 & 40.25 & 8.00 & \underline{91.81} & \textbf{66.33} & 62.17 & 47.85 \\
           & 0.5 & 28.34 & 39.54 & \textbf{51.11} & \underline{61.22} & \textbf{46.78} & 30.13 & \underline{33.38} & 24.00 & \textbf{24.94} & 64.00 & 86.96 & 39.81 & 8.50 & 87.46 & \textbf{65.41} & 62.30 & 47.12 \\
           & 0.75 & 27.23 & 34.65 & 39.60 & \underline{57.52} & \underline{39.27} & 27.58 & \underline{32.04} & 22.76 & \underline{23.70} & \underline{70.00} & \underline{85.29} & \textbf{39.71} & \textbf{10.00} & 49.74 & \textbf{59.30} & \underline{63.46} & 42.62 \\
           & 0.9 & 23.28 & \underline{27.76} & 32.42 & 43.62 & \underline{31.74} & 23.28 & \underline{29.65} & 20.98 & \underline{22.00} & \underline{61.75} & 84.16 & \textbf{38.15} & \textbf{11.50} & 17.42 & \textbf{50.93} & \underline{62.92} & 36.35 \\
          \midrule
          \multirow{4}{*}{keydiff} & 0.25 & 25.44 & 35.02 & 49.12 & 47.28 & 42.49 & 23.07 & 32.84 & 22.94 & 24.93 & 62.50 & 83.61 & 38.84 & 6.37 & 91.50 & 61.75 & 56.71 & 44.03 \\
           & 0.5 & 23.30 & 29.97 & \underline{39.86} & 35.86 & 33.86 & 16.60 & 29.49 & 22.43 & 21.93 & 55.00 & 85.62 & 38.25 & 5.49 & 86.33 & 46.51 & 53.65 & 39.01 \\
           & 0.75 & 14.90 & 18.21 & 29.26 & 19.68 & 27.49 & 9.51 & 21.97 & 21.18 & 14.22 & 39.00 & 81.56 & 36.15 & 5.81 & 45.58 & 27.38 & 53.66 & 29.10 \\
           & 0.9 & 10.62 & 7.80 & 23.18 & 15.76 & 23.01 & 7.01 & 16.07 & 19.85 & 8.02 & 6.50 & 78.15 & 31.41 & 7.78 & 13.92 & 13.57 & 52.57 & 20.95 \\
          \midrule
          \multirow{4}{*}{knorm} & 0.25 & 25.35 & 34.04 & 49.59 & 49.24 & 40.66 & 24.62 & 32.53 & 23.17 & 24.51 & 63.00 & 84.55 & \textbf{41.94} & \underline{11.50} & 93.00 & 50.21 & 47.71 & 43.48 \\
           & 0.5 & 17.84 & 24.04 & 39.55 & 29.63 & 28.61 & 14.82 & 28.85 & 21.55 & 21.26 & 50.50 & 81.30 & \textbf{40.80} & 6.64 & 78.46 & 36.13 & 51.59 & 35.72 \\
           & 0.75 & 13.39 & 12.33 & 27.76 & 13.90 & 18.25 & 6.39 & 22.63 & 20.24 & 14.73 & 25.75 & 81.36 & 38.51 & 9.88 & 20.00 & 16.83 & 54.66 & 24.79 \\
           & 0.9 & 9.66 & 7.65 & 23.48 & 8.78 & 19.54 & 4.11 & 15.46 & 19.24 & 8.15 & 14.00 & 80.68 & 33.08 & 3.75 & 6.50 & 11.54 & 55.83 & 20.09 \\
           \midrule
          \multirow{4}{*}{\shortstack{Jacap \\ (ours)}} & 0.25 & \underline{28.61} & \underline{43.45} & \underline{53.05} & \underline{62.37} & 48.04 & 33.94 & \textbf{33.84} & 24.09 & 24.72 & \underline{71.50} & \textbf{90.01} & 40.63 & \textbf{12.50} & \textbf{95.54} & 65.10 & \underline{63.02} & \underline{49.40} \\
           & 0.5 & \underline{28.40} & \textbf{43.92} & 50.90 & \textbf{62.61} & \underline{46.25} & \underline{34.04} & \textbf{33.48} & \textbf{24.47} & \underline{24.87} & \textbf{75.00} & \textbf{87.82} & \underline{40.01} & \underline{9.50} & \underline{96.71} & \underline{62.24} & \underline{62.91} & \textbf{48.95} \\
           & 0.75 & \underline{27.70} & \textbf{39.07} & \underline{43.48} & \textbf{59.54} & \textbf{42.13} & \underline{28.50} & \textbf{32.60} & \underline{23.44} & \textbf{23.89} & \textbf{72.50} & 84.36 & \underline{39.55} & \textbf{10.00} & \underline{95.96} & \underline{57.65} & 63.36 & \textbf{46.48} \\
           & 0.9 & \underline{24.12} & \textbf{31.39} & \underline{34.32} & \textbf{49.68} & \textbf{37.26} & \textbf{28.34} & \textbf{30.21} & \textbf{22.80} & \textbf{22.14} & \textbf{62.25} & \textbf{86.90} & \underline{36.56} & \underline{9.00} & \textbf{70.29} & \underline{46.93} & 62.37 & \textbf{40.91} \\
            \midrule

          \multicolumn{2}{l}{\textbf{Qwen3-14B}} & 30.32 & 44.05 & 52.14 & 62.30 & 55.87 & 32.79 & 32.84 & 24.32 & 24.96 & 70.00 & 89.10 & 42.01 & 9.80 & 99.42 & 69.36 & 67.27 & 50.41 \\
          \midrule
          \multirow{4}{*}{CapKV} & 0.25 & \textbf{30.56} & 43.25 & \textbf{52.49} & \textbf{63.55} & 54.25 & \textbf{35.68} & 32.85 & \textbf{24.66} & 24.88 & \textbf{75.00} & \underline{90.10} & \underline{41.80} & 8.00 & \underline{99.08} & \textbf{69.42} & \textbf{67.25} & \textbf{50.80} \\
           & 0.5 & \textbf{29.92} & \textbf{43.96} & \textbf{51.95} & \textbf{65.06} & \underline{54.11} & \textbf{35.42} & \underline{32.80} & \underline{24.19} & 24.87 & \underline{72.00} & \textbf{91.85} & 41.27 & 9.50 & \textbf{97.62} & 66.82 & \textbf{67.84} & \textbf{50.57} \\
           & 0.75 & \underline{27.62} & \underline{37.53} & \underline{45.64} & \textbf{58.93} & \underline{46.67} & \underline{32.22} & 31.09 & \underline{24.24} & 23.70 & 66.00 & \underline{89.71} & 40.80 & 7.15 & \textbf{98.58} & 57.86 & \underline{66.89} & \underline{47.16} \\
           & 0.9 & \underline{25.16} & \underline{27.78} & \underline{35.16} & \underline{49.75} & \underline{32.62} & \underline{26.88} & 27.45 & \underline{21.98} & 20.70 & 48.25 & 87.82 & 38.41 & 5.50 & \underline{76.67} & 45.08 & \underline{63.25} & \underline{39.53} \\
          \midrule
          \multirow{4}{*}{EA} & 0.25 & \underline{29.31} & \textbf{44.25} & 51.06 & \underline{63.10} & \textbf{56.44} & 32.64 & 32.79 & 24.47 & 24.85 & 70.50 & 89.60 & 41.49 & 9.00 & 97.45 & \underline{69.30} & 66.54 & 50.17 \\
           & 0.5 & 28.28 & \underline{43.70} & 49.47 & \underline{63.83} & 50.77 & 31.47 & 32.29 & 24.13 & \underline{25.15} & 70.50 & \underline{90.56} & \underline{41.46} & 8.70 & 94.17 & \textbf{67.75} & 65.82 & 49.25 \\
           & 0.75 & 23.62 & 35.92 & 37.70 & 52.92 & 38.91 & 26.90 & \underline{31.37} & 22.79 & \underline{24.74} & \underline{66.50} & \textbf{90.10} & \textbf{41.20} & 8.55 & 72.74 & \underline{61.32} & 64.21 & 43.72 \\
           & 0.9 & 22.78 & 25.16 & 27.36 & 41.20 & 23.29 & 20.64 & \underline{28.79} & 21.28 & \underline{22.39} & \underline{61.42} & \textbf{89.02} & \textbf{39.65} & \textbf{10.50} & 27.96 & \underline{50.87} & 62.00 & 35.89 \\
          
          \midrule
          \multirow{4}{*}{keydiff} & 0.25 & 28.11 & 36.18 & 50.24 & 51.61 & 45.09 & 26.48 & \underline{33.02} & 24.06 & 24.94 & 69.00 & 82.57 & 41.52 & \textbf{11.50} & 98.58 & 66.02 & 62.22 & 46.95 \\
           & 0.5 & 24.93 & 30.20 & 44.89 & 41.04 & 35.61 & 21.60 & 28.69 & 23.16 & 22.65 & 57.50 & 75.73 & 39.56 & \textbf{13.00} & 95.50 & 49.69 & 57.42 & 41.32 \\
           & 0.75 & 21.03 & 21.80 & 34.10 & 26.79 & 27.78 & 14.63 & 20.31 & 21.41 & 15.55 & 33.00 & 74.33 & 34.99 & \underline{10.50} & 81.83 & 28.68 & 54.69 & 32.59 \\
           & 0.9 & 12.76 & 12.12 & 24.13 & 18.97 & 25.03 & 6.77 & 11.41 & 19.50 & 6.68 & 3.00 & 74.32 & 30.10 & \textbf{10.50} & 39.00 & 15.19 & 54.80 & 22.77 \\
          \midrule
          \multirow{4}{*}{knorm} & 0.25 & 28.28 & 40.58 & \underline{52.17} & 54.61 & 43.52 & 28.23 & \textbf{33.24} & 23.74 & \underline{24.95} & 65.00 & 79.56 & 39.75 & \underline{10.50} & \textbf{100.00} & 58.96 & 62.29 & 46.59 \\
           & 0.5 & 23.58 & 34.56 & 44.65 & 43.17 & 36.84 & 15.95 & 26.18 & 22.96 & 22.80 & 60.00 & 78.35 & 39.15 & \underline{12.50} & \underline{95.08} & 44.16 & 59.78 & 41.23 \\
           & 0.75 & 17.72 & 17.80 & 35.08 & 22.71 & 25.99 & 6.97 & 15.05 & 20.99 & 16.87 & 44.50 & 82.07 & 36.27 & \textbf{13.00} & 63.58 & 27.58 & 54.15 & 31.27 \\
           & 0.9 & 10.07 & 16.59 & 26.61 & 11.10 & 21.02 & 4.54 & 10.45 & 19.66 & 11.37 & 23.50 & 77.52 & 33.51 & \underline{8.50} & 13.58 & 20.85 & 46.55 & 22.21 \\

           \midrule
          \multirow{4}{*}{\shortstack{Jacap \\ (ours)}} & 0.25 & 29.28 & \underline{43.70} & 52.14 & 62.43 & \underline{56.03} & \underline{35.50} & 32.96 & \underline{24.50} & \textbf{25.13} & \underline{71.00} & \textbf{90.85} & \textbf{41.95} & 9.00 & 97.75 & 69.24 & \underline{67.20} & \underline{50.54} \\
           & 0.5 & \underline{28.50} & 42.62 & \underline{51.19} & 63.26 & \textbf{54.65} & \underline{34.06} & \textbf{32.81} & \textbf{24.79} & \textbf{25.22} & \textbf{74.00} & 90.31 & \textbf{42.27} & 7.05 & \textbf{97.62} & \underline{67.43} & \underline{67.32} & \underline{50.19} \\
           & 0.75 & \textbf{27.90} & \textbf{41.32} & \textbf{45.78} & \underline{57.85} & \textbf{52.86} & \textbf{33.29} & \textbf{32.09} & \textbf{24.83} & \textbf{24.86} & \textbf{75.50} & 88.94 & \underline{41.08} & 7.99 & \underline{96.92} & \textbf{62.58} & \textbf{67.92} & \textbf{48.86} \\
           & 0.9 & \textbf{27.19} & \textbf{32.42} & \textbf{38.16} & \textbf{54.21} & \textbf{45.41} & \textbf{30.60} & \textbf{29.59} & \textbf{22.62} & \textbf{23.21} & \textbf{68.50} & \underline{88.57} & \underline{38.62} & 8.56 & \textbf{88.08} & \textbf{51.73} & \textbf{66.00} & \textbf{44.59} \\
            \midrule
          \multicolumn{2}{l}{\textbf{Llama3.1-8B}} & 30.46 & 47.27 & 55.98 & 59.00 & 51.23 & 33.55 & 35.24 & 25.27 & 26.80 & 29.50 & 86.01 & 39.31 & 10.65 & 100.00 & 53.47 & 47.73 & 45.72 \\
            \midrule
          \multirow{4}{*}{CapKV} & 0.25 & \textbf{31.53} & \textbf{47.74} & \textbf{56.95} & 57.31 & \textbf{51.90} & 32.91 & 34.61 & 24.65 & 26.78 & \textbf{68.50} & \textbf{90.21} & 38.80 & \textbf{13.65} & \textbf{100.00} & \underline{53.47} & \underline{48.52} & \textbf{48.60} \\
           & 0.5 & \underline{31.69} & 44.84 & 51.42 & \textbf{58.24} & 48.35 & \underline{30.66} & \textbf{33.88} & 24.65 & 26.47 & \underline{65.25} & \textbf{90.71} & 37.72 & \underline{12.00} & 98.50 & 50.12 & \underline{49.47} & \underline{47.12} \\
           & 0.75 & 29.73 & \underline{39.93} & \underline{43.71} & 53.19 & \textbf{46.49} & 24.66 & 29.32 & \underline{23.71} & 25.19 & 30.75 & \underline{90.83} & 33.44 & 9.00 & 66.50 & 44.47 & \textbf{50.91} & 40.11 \\
           & 0.9 & 25.33 & 24.72 & 27.81 & 43.90 & 30.86 & 21.58 & 24.56 & 22.14 & 21.54 & 7.00 & \underline{91.22} & 28.79 & \underline{10.00} & 21.50 & 35.61 & 51.06 & 30.48 \\
          \midrule
          \multirow{4}{*}{EA} & 0.25 & 31.17 & 47.49 & \underline{56.85} & 58.33 & 49.98 & 33.82 & \underline{34.94} & 24.78 & \underline{27.18} & 29.50 & 85.66 & 38.03 & \underline{11.20} & \underline{99.00} & 52.49 & 47.54 & 45.50 \\
           & 0.5 & 31.19 & \underline{46.67} & 51.49 & \underline{56.34} & \underline{48.65} & \textbf{33.48} & 33.63 & \underline{24.74} & \underline{27.04} & 19.00 & 85.91 & 38.87 & \textbf{12.50} & 87.00 & \textbf{52.97} & 48.64 & 43.63 \\
           & 0.75 & \underline{30.17} & 39.70 & 41.55 & \textbf{55.53} & 44.55 & \underline{26.75} & \textbf{31.77} & 23.26 & \textbf{26.06} & 27.00 & 88.98 & \underline{38.59} & \textbf{10.00} & 45.00 & \textbf{49.84} & 50.01 & 39.30 \\
           & 0.9 & 25.87 & \underline{29.04} & 32.51 & \underline{48.46} & \underline{31.27} & \underline{24.48} & \textbf{29.07} & \textbf{22.72} & \textbf{24.60} & \textbf{50.50} & \textbf{92.17} & 26.90 & 7.55 & 18.00 & \underline{40.11} & 48.94 & 34.51 \\
          \midrule
          \multirow{4}{*}{keydiff} & 0.25 & \underline{31.50} & 47.20 & 54.17 & \textbf{58.79} & \underline{50.95} & \textbf{35.85} & 34.48 & 24.77 & 26.83 & 62.50 & 86.56 & \textbf{40.13} & 10.65 & 99.50 & 52.25 & 47.03 & \underline{47.70} \\
           & 0.5 & \textbf{32.30} & 44.93 & \textbf{52.98} & 55.52 & 46.60 & 30.26 & 32.45 & 24.47 & 25.71 & 64.00 & 87.29 & \underline{40.35} & 10.65 & \textbf{99.50} & 45.18 & 46.77 & 46.18 \\
           & 0.75 & 29.62 & 33.01 & 41.37 & 52.51 & 37.97 & 25.28 & 29.41 & \textbf{23.95} & 22.93 & \underline{52.00} & 84.36 & \textbf{40.03} & \underline{9.66} & \textbf{99.50} & 37.48 & 47.58 & \underline{41.67} \\
           & 0.9 & \underline{28.49} & 16.95 & \underline{34.22} & 40.71 & 25.38 & 15.27 & 26.30 & 21.03 & 19.15 & 40.00 & 80.01 & \textbf{39.18} & 8.50 & \textbf{89.00} & 26.07 & 47.99 & \underline{34.89} \\
          \midrule
          \multirow{4}{*}{knorm} & 0.25 & 30.71 & 45.34 & 56.16 & \underline{58.43} & 50.63 & \underline{35.27} & 34.07 & \underline{24.83} & 26.67 & \underline{66.00} & 87.26 & 36.85 & 11.15 & 97.50 & 36.91 & 48.35 & 46.63 \\
           & 0.5 & 27.45 & 39.59 & 49.46 & 55.72 & 44.23 & 26.95 & 32.22 & 24.25 & 25.51 & 59.00 & 86.44 & 33.60 & 9.55 & 87.50 & 33.05 & 49.16 & 42.73 \\
           & 0.75 & 22.18 & 29.67 & 39.44 & 47.45 & 30.45 & 20.51 & 28.68 & 22.78 & 23.48 & 51.50 & 72.85 & 28.98 & 9.15 & 52.50 & 27.20 & 49.17 & 34.75 \\
           & 0.9 & 22.44 & 16.23 & 28.48 & 37.69 & 21.73 & 13.30 & 26.03 & 20.64 & 20.56 & \underline{45.50} & 62.78 & \underline{29.03} & \textbf{11.50} & 25.00 & 23.06 & \textbf{51.52} & 28.47 \\
           \midrule
          \multirow{4}{*}{\shortstack{Jacap \\ (ours)}} & 0.25 & 31.46 & \underline{47.52} & 55.57 & 57.79 & 50.26 & 32.54 & \textbf{35.04} & \textbf{24.99} & \textbf{27.30} & 44.50 & \underline{90.06} & \underline{39.82} & 11.15 & \textbf{100.00} & \textbf{53.68} & \textbf{50.50} & 47.01 \\
           & 0.5 & 30.32 & \textbf{46.74} & \underline{52.37} & 55.51 & \textbf{48.92} & 27.33 & \underline{33.71} & \textbf{24.75} & \textbf{27.14} & \textbf{66.00} & \underline{89.31} & \textbf{40.59} & 11.00 & \underline{99.00} & \underline{51.78} & \textbf{49.87} & \textbf{47.15} \\
           & 0.75 & \textbf{30.77} & \textbf{42.32} & \textbf{45.32} & \underline{54.92} & \underline{45.89} & \textbf{28.65} & \underline{31.05} & 23.34 & \underline{25.94} & \textbf{56.50} & \textbf{91.96} & 37.26 & 7.00 & \underline{87.50} & \underline{48.04} & \underline{50.51} & \textbf{44.19} \\
           & 0.9 & \textbf{29.13} & \textbf{30.94} & \textbf{35.46} & \textbf{49.26} & \textbf{34.00} & \textbf{25.15} & \underline{27.52} & \underline{22.64} & \underline{24.48} & 32.50 & 90.40 & 27.93 & 8.00 & \underline{46.50} & \textbf{40.52} & \underline{51.21} & \textbf{35.98} \\
          \bottomrule
        \end{tabular}
      \end{sc}
    \end{scriptsize}
  \end{center}
  \label{tab:longbench_results}
  \vskip -0.1in
\end{table}

We evaluate \textsc{Jacap} against \textsc{CapKV}, \textsc{SnapKV}, \textsc{KeyDiff}, \textsc{EA}, and \textsc{KNorm} on the LongBench~\cite{bai2023longbench} across four compression ratios ($0.25$, $0.5$, $0.75$, $0.9$), using Qwen3-8B, Qwen3-14B~\cite{yang2025qwen3}, and Llama3.1-8B~\cite{grattafiori2024llama}. Table~\ref{tab:longbench_results} summarizes the results. We focus on high compression ratios (0.75 and 0.9), where the memory constraints are most severe and differences among methods are pronounced.

At 0.75 compression, \textsc{Jacap} generally matches or slightly exceeds \textsc{CapKV} on average, while significantly outperforming heuristic baselines (\textsc{SnapKV}, \textsc{KeyDiff}, \textsc{EA}, \textsc{KNorm}). When compression is further increased to 0.9, \textsc{Jacap} achieves the largest advantage, leading the average score by approximately 5 points over the next best baseline. This demonstrates its superior ability to retain critical KV information under extreme cache constraints.

At the task level, \textsc{Jacap} maintains stronger performance on both SD-QA and MD-QA tasks, indicating better preservation of context-relevant information for long-context reasoning. It also exhibits more stable degradation on summarization and few-shot tasks as compression increases. These trends remain consistent across all three models, demonstrating the robustness and generality of the proposed method.

\subsection{Experiments on NIAH}

We evaluate the performance of \textsc{Jacap} and baseline KV-cache eviction methods on the Needle-in-Haystack (NIAH)~\cite{niah}benchmark under a compression ratio of 0.75 using Qwen3-8B. The NIAH setup tests extremely long contexts (10k–120k tokens) and varying Needle Depths (0.15–0.95), simulating sparse target retrieval under constrained cache budgets.

Figure~\ref{fig:niah_75} presents the retrieval performance heatmaps across different context lengths and Needle Depths. \textsc{Jacap} consistently achieves stronger retrieval performance, particularly for deep needles and long contexts where heuristic baselines show significant degradation. Although \textsc{CapKV} performs competitively at moderate depths, \textsc{Jacap} maintains more stable performance across the full range of settings. These results suggest that Jacobian-capacity-aware eviction more effectively preserves KV pairs relevant to sparse and distant queries, leading to more reliable retrieval in extreme long-context scenarios. Additional results are provided in Appendix~\ref{app:mer_niah}.

\subsection{Experiments on Decoding Eviction}
\vspace{-10pt}
\begin{table}[htbp]
\centering
\caption{Performance comparison of different KV-cache eviction methods on AIME25 under decoding-stage eviction using Nemotron-7B. The number of reserved tokens controls the decoding cache budget. \textbf{Bold} indicates the best-performing method under each budget.}
\vspace{10pt}
\renewcommand{\arraystretch}{1.3} 
\setlength{\tabcolsep}{6pt}       
\begin{tabular}{c|lllll}
\hline
\textbf{Reserved Tokens} & \textbf{EA} & \textbf{KeyDiff} & \textbf{KNorm} & \textbf{CapKV} & \textbf{Jacap (Ours)} \\ 
\hline
2048  & 0.17 & 0.07 & 0     & 0.20 & \textbf{0.30} \\ 
4096  & 0.30 & 0.33 & 0.03  & 0.50 & 0.50          \\ 
8192  & 0.40 & 0.53 & 0.27  & \textbf{0.73} & 0.53          \\ 
16384 & 0.63 & 0.67 & 0.57  & 0.67 & \textbf{0.73} \\ 
\hline
\end{tabular}
\label{tab:decoding_eviction}
\end{table}
We evaluate \textsc{Jacap} during online decoding on AIME25~\cite{aime25} using Nemotron-7B~\cite{ahmad2025opencodereasoning}. Table~\ref{tab:decoding_eviction} summarizes performance across different reserved token budgets, compared with \textsc{EA}, \textsc{KeyDiff}, \textsc{KNorm}, and \textsc{CapKV}. At the smallest cache budget (2048 tokens), \textsc{Jacap} achieves the best score, showing its effectiveness in preserving critical KV entries under extreme constraints. With 4096 reserved tokens, it matches \textsc{CapKV} and clearly outperforms other baselines. At 8192 tokens, \textsc{CapKV} surpasses \textsc{Jacap}, but when the budget increases to 16384 tokens, \textsc{Jacap} recovers the lead, outperforming all methods.

These results indicate that \textsc{Jacap} remains competitive in dynamic decoding scenarios, especially under tight or large cache budgets. Its sensitivity-aware, non-redundant selection helps maintain reasoning performance throughout autoregressive generation, highlighting the practical benefits of Jacobian-capacity-aware eviction.

\subsection{Ablation Study}

\begin{wrapfigure}{r}[-20pt]{0.4\linewidth}
  \centering
  \includegraphics[width=\linewidth]{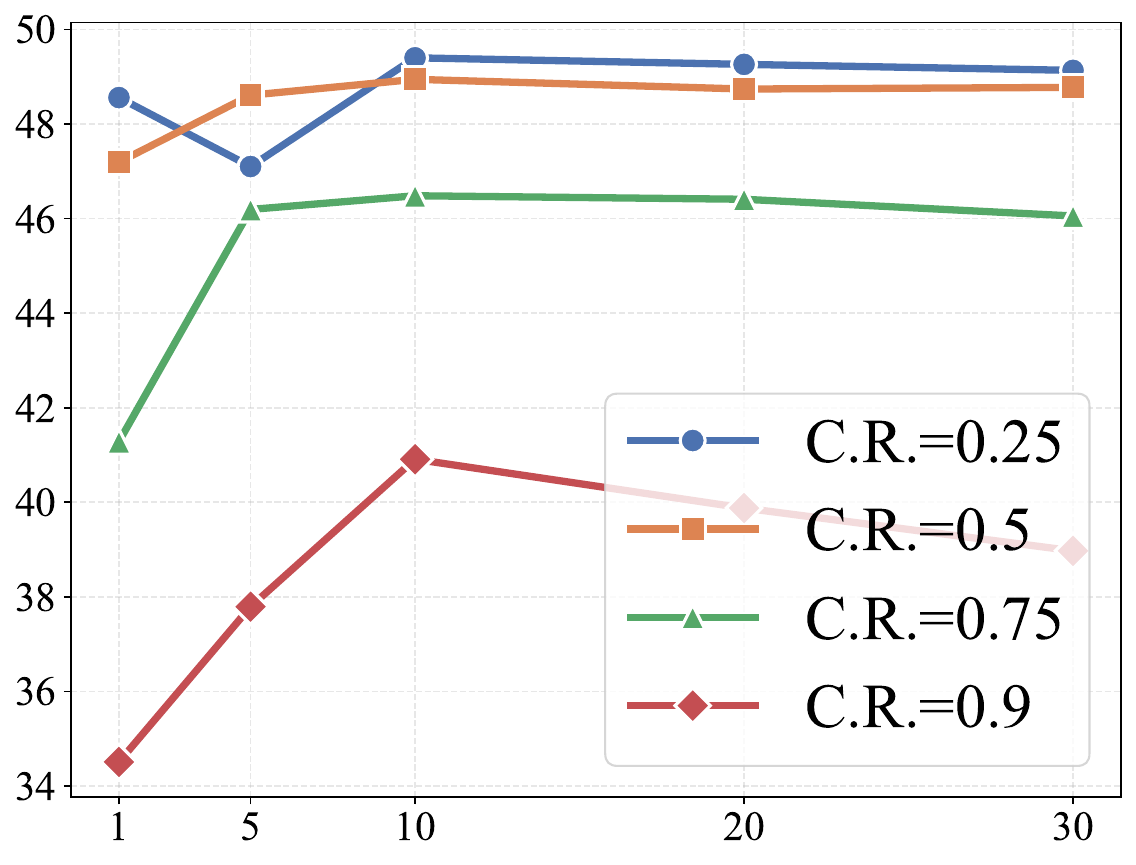}
  \caption{Ablation study of the temperature parameter $\tau$ on LongBench.}
  \label{fig:ablation_tau}
\end{wrapfigure}

Figure~\ref{fig:ablation_tau} reports the average LongBench performance of Qwen3-8B under different temperature values and compression ratios. We observe that \textsc{Jacap} achieves the best overall performance when $\tau$ is around $10$. A small $\tau$ makes the estimated attention prior overly sharp, causing the eviction score to focus excessively on a few high-alignment tokens and potentially discard other informative entries. Conversely, a large $\tau$ over-smooths the prior, weakening the query-dependent relevance signal and reducing the method toward a more diversity-dominated selection rule. The optimal performance around $\tau=10$ indicates that a moderate temperature provides a suitable balance between local query relevance and cache diversity.
\vspace{-5pt}
\section{Conclusion}
\vspace{-10pt}

\label{sec:conclusion}

In this work, we introduced a local nonlinear information-theoretic framework to analyze the predictive capacity of KV caches in LLMs, explicitly capturing the competition and sensitivity induced by softmax attention. Based on this framework, we proposed \textsc{Jacap}, a Jacobian-capacity-aware eviction strategy that prioritizes informative and non-redundant tokens. Experiments on standard reasoning benchmarks, extreme long-context retrieval, and dynamic decoding tasks show that \textsc{Jacap} outperforms existing heuristics and capacity-aware baselines, especially under tight cache constraints. Nevertheless, our method relies on a first-order local approximation and simplifies pairwise softmax competition into token-wise sensitivity weights, which may limit its expressiveness in highly dynamic decoding scenarios. Future work may explore higher-order local approximations, more faithful competition modeling, and adaptive cache management across layers and decoding stages.
\bibliographystyle{plainnat}
\bibliography{references}


\appendix

\section{Derivation and analysis of Jacobian Information Capacity}
\label{sec:ana_jic}

In this appendix, we provide the rigorous mathematical derivation of the Jacobian Information Capacity presented in Theorem 1 of the main text. We begin by explicitly defining the local linearization framework and then derive the closed-form mutual information objective under the resulting Gaussian approximation. Finally, we analyze the structure of the derived Jacobian matrix to provide deeper insights into how it captures the nonlinear dynamics of softmax attention.

\subsection{Proof of Theorem ~\ref{theorem:jacobian_logdet}}
\label{app:proof_theorem1}
Let $\mathcal Z_{\mathcal C} = \{(k_i, v_i)\}_{i \in \mathcal{C}}$ be the retained subset of key-value pairs, where $|\mathcal{C}|=m$. Let $K_\mathcal{C} \in \mathbb{R}^{m \times d_k}$ and $U_\mathcal{C} \in \mathbb{R}^{d_o \times m}$ be the stacked key and output-value matrices, respectively. The retained attention mechanism defines a nonlinear mapping $f_\mathcal{C}: \mathbb{R}^{d_k} \to \mathbb{R}^{d_o}$ from a query $q$ to the output $y$:
$$
f_\mathcal{C}(q) = U_\mathcal{C} \operatorname{softmax}\left(\frac{K_\mathcal{C} q}{\sqrt{d_k}}\right).
$$
    
We model the attention process as a noisy nonlinear channel:
$$
Y_\mathcal{C} = f_\mathcal{C}(q) + \epsilon, \quad \epsilon \sim \mathcal{N}(0, \Sigma_{\text{noise}}).
$$

We assume future queries are localized around a center $\mu_Q$ with covariance $\Lambda_Q$, i.e., $q \sim \mathcal{N}(\mu_Q, \Lambda_Q)$. Our goal is to approximate the mutual information $I(q; Y_\mathcal{C} | \mathcal Z_{\mathcal C})$.

To make the analysis tractable, we perform a first-order Taylor expansion of the nonlinear mapping $f_\mathcal{C}(q)$ around the query center $\mu_Q$. Let $\delta q = q - \mu_Q$. The expansion is given by:
$$
    f_\mathcal{C}(q) \approx f_\mathcal{C}(\mu_Q) + J_\mathcal{C} \delta q,
$$
where $J_\mathcal{C} = \left. \frac{\partial f_\mathcal{C}(q)}{\partial q} \right|_{q=\mu_Q}$ is the Jacobian matrix evaluated at $\mu_Q$.

We now derive the explicit form of $J_\mathcal{C}$. Let $s(q) = \frac{1}{\sqrt{d_k}} K_\mathcal{C} q \in \mathbb{R}^m$ be the logits, and $\alpha(s) = \operatorname{softmax}(s) \in \mathbb{R}^m$ be the attention weights. By the chain rule:
$$
    \frac{\partial f_\mathcal{C}}{\partial q} = U_\mathcal{C} \frac{\partial \alpha}{\partial s} \frac{\partial s}{\partial q}.
$$
The derivative of the logits with respect to $q$ is $\frac{\partial s}{\partial q} = \frac{1}{\sqrt{d_k}} K_\mathcal{C}$. The derivative of the softmax function $\alpha(s)$ with respect to its input logits $s$ is given by the standard result:
$$
    \frac{\partial \alpha}{\partial s} = \operatorname{Diag}(\alpha) - \alpha \alpha^\top.
$$
Evaluating these derivatives at $q = \mu_Q$, we define the attention weights at the center as $\alpha_\mathcal{C}^\star = \operatorname{softmax}(\frac{K_\mathcal{C} \mu_Q}{\sqrt{d_k}})$, and the softmax Jacobian matrix as $S_\mathcal{C} = \operatorname{Diag}(\alpha_\mathcal{C}^\star) - \alpha_\mathcal{C}^\star (\alpha_\mathcal{C}^\star)^\top$. Substituting these back yields the Jacobian of the attention map:
\begin{equation}
    J_\mathcal{C} = \frac{1}{\sqrt{d_k}} U_\mathcal{C} S_\mathcal{C} K_\mathcal{C}.
\end{equation}
This completes the proof of Lemma~\ref{lemma:local_linear} and provides the explicit form of the effective local linear channel.

Under the first-order approximation, the noisy channel becomes a linear Gaussian channel with respect to the perturbation $\delta q \sim \mathcal{N}(0, \Lambda_Q)$:
\begin{equation}
    Y_\mathcal{C} \approx \underbrace{f_\mathcal{C}(\mu_Q)}_{\text{constant bias}} + J_\mathcal{C} \delta q + \epsilon.
\end{equation}
Since $\delta q$ and $\epsilon$ are independent Gaussian random variables, the output $Y_\mathcal{C}$ is also Gaussian. The mutual information is invariant to the constant bias term. We compute the relevant covariance matrices:

\begin{equation}
    \begin{split}
        \Sigma_{Y|q} & = \operatorname{Cov}(Y_\mathcal{C} | q) = \operatorname{Cov}(\epsilon) = \Sigma_{\text{noise}}.\\
        \Sigma_Y & = \operatorname{Cov}(Y_\mathcal{C}) = J_\mathcal{C} \operatorname{Cov}(\delta q) J_\mathcal{C}^\top + \operatorname{Cov}(\epsilon) = J_\mathcal{C} \Lambda_Q J_\mathcal{C}^\top + \Sigma_{\text{noise}}.
    \end{split}
\end{equation}

For Gaussian distributions, the mutual information is the difference between marginal and conditional differential entropies, which can be calculated via determinants of covariance matrices:
\begin{equation}
    \begin{split}
    \mathcal{L}_\mathcal{C}^{\text{Jac}} \approx I(q; Y_\mathcal{C} | \mathcal Z_{\mathcal C}) &= H(Y_\mathcal{C}) - H(Y_\mathcal{C} | q) \\
    &= \frac{1}{2} \log \det(\Sigma_Y) - \frac{1}{2} \log \det(\Sigma_{Y|q}) \\
    &= \frac{1}{2} \log \frac{\det(J_\mathcal{C} \Lambda_Q J_\mathcal{C}^\top + \Sigma_{\text{noise}})}{\det(\Sigma_{\text{noise}})}.
    \end{split}
\end{equation}
Applying the matrix determinant identity $\det(A+B) = \det(A)\det(I + A^{-1}B)$, we obtain the final closed-form objective:
$$
    \mathcal{L}_\mathcal{C}^{\text{Jac}} \approx \frac{1}{2} \log \det \left( I + \Sigma_{\text{noise}}^{-1} J_\mathcal{C} \Lambda_Q J_\mathcal{C}^\top \right).
$$
This completes the derivation of Theorem ~\ref{theorem:jacobian_logdet}.

\subsection{Analysis of the Jacobian Information Capacity}
\label{app:jacobian_ana}

The derived Jacobian capacity differs from prior linear information-theoretic objectives primarily through the structure of the Jacobian matrix $J_\mathcal{C} \propto U_\mathcal{C} S_\mathcal{C} K_\mathcal{C}$. The core novelty lies in the \textbf{Softmax Competition Matrix} $S_\mathcal{C} = \operatorname{Diag}(\alpha_\mathcal{C}^\star) - \alpha_\mathcal{C}^\star (\alpha_\mathcal{C}^\star)^\top$, which explicitly captures the nonlinear dynamics of attention. Here we analyze its two key components:

\paragraph{1. Local Sensitivity and Saturation.}
The diagonal elements of $S_\mathcal{C}$ are given by $S_{ii} = \alpha_i^\star(1 - \alpha_i^\star)$, where $\alpha_i^\star$ is the attention weight of the $i$-th token at the query center $\mu_Q$. This term measures the local sensitivity of the softmax function. $S_\mathcal{C}$ offers two new and important insights:
\begin{itemize}
    \item \textbf{Peak Sensitivity:} The term is maximized when $\alpha_i^\star \approx 0.5$, indicating that tokens with moderate attention weights are most sensitive to query perturbations and thus provide the highest potential information gain locally.
    \item \textbf{Saturation Regions:} As $\alpha_i^\star \to 0$ (irrelevant) or $\alpha_i^\star \to 1$ (dominant), the term $S_{ii} \to 0$. This captures the saturation effect: if a token is already ignored or fully dominant, small changes in the query will not significantly change its contribution to the output.
\end{itemize}
This contrasts sharply with linear models or heuristics that assume higher attention weights always imply higher importance. Our theory suggests that saturated tokens, despite high attention, may have low marginal information value.

\paragraph{2. Token Competition.}
The off-diagonal elements $S_{ij} = -\alpha_i^\star \alpha_j^\star$ (for $i \neq j$) are always negative. This explicitly models the competition introduced by softmax normalization. An increase in the logit of token $j$ necessarily decreases the attention weights of all other tokens $i$, coupling their contributions.

The linear gaussian surrogate~\cite{yang2026rethinkingkvcacheeviction} effectively assumes $S_\mathcal{C}$ is proportional to the identity matrix. By explicitly incorporating $S_\mathcal{C}$, the Jacobian Information Capacity rewards retained subsets that are not only aligned with future queries ($K_\mathcal{C}$ and $\alpha_\mathcal{C}^\star$) but also possess high local sensitivity and distinct output directions ($U_\mathcal{C}$) under the constraints of softmax competition.

\section{More details on Jacap Eviction Method}
\label{app:ana_jacap}
In this appendix, we detail the principled approximations that reduce the full Jacobian Information Capacity to the efficient, one-shot \textsc{Jacap} scoring rule presented in Algorithm~\ref{alg:jacap}. We also provide a complexity analysis of the proposed algorithm.

\subsection{Derivation of the \textsc{Jacap} Algorithm}
\label{app:jacap_der}
The complexity of the objective $\mathcal{L}_\mathcal{C}^{\text{Jac}} \propto \log \det (I + \Sigma_{\text{noise}}^{-1} U_\mathcal{C} S_\mathcal{C} K_\mathcal{C} \Lambda_Q K_\mathcal{C}^\top S_\mathcal{C}^\top U_\mathcal{C}^\top / d_k)$ stems from three coupled structured matrices: the softmax competition matrix $S_\mathcal{C}$, the query-key response covariance $K_\mathcal{C} \Lambda_Q K_\mathcal{C}^\top$, and the subset-dependent selection problem itself. We address these via a series of approximations.

\paragraph{Decoupling Softmax Competition via Diagonal Approximation.}

The matrix $S_\mathcal{C} = \operatorname{Diag}(\alpha_\mathcal{C}^\star) - \alpha_\mathcal{C}^\star (\alpha_\mathcal{C}^\star)^\top$ couples all tokens through its off-diagonal terms. To enable efficient token-wise scoring, we approximate $S_\mathcal{C}$ by retaining only its diagonal elements, which capture the local sensitivity of each token's weight to its own logit:
$$
    S_\mathcal{C} \approx D_\mathcal{C} := \operatorname{Diag}(\rho_i)_{i \in \mathcal{C}}, \quad \text{where } \rho_i = S_{ii} = \alpha_i^\star(1 - \alpha_i^\star).
$$

As analyzed in Appendix~\ref{app:jacobian_ana}, the diagonal term $\rho_i$ perfectly encapsulates the "sensitivity vs. saturation" trade-off. While the off-diagonal terms $-\alpha_i^\star \alpha_j^\star$ model negative pairwise competition, neglecting them is a common and effective simplification in variational inference and large-scale approximations. It preserves the most critical nonlinear effect—that saturated tokens have low local influence—while decoupling token interactions in the sensitivity term.

Under this approximation, the Jacobian becomes $J_\mathcal{C} \approx \frac{1}{\sqrt{d_k}} U_\mathcal{C} D_\mathcal{C} K_\mathcal{C}$, and the effective channel matrix inside the log-determinant becomes:
$$
    J_\mathcal{C} \Lambda_Q J_\mathcal{C}^\top \approx \frac{1}{d_k} U_\mathcal{C} D_\mathcal{C} (K_\mathcal{C} \Lambda_Q K_\mathcal{C}^\top) D_\mathcal{C} U_\mathcal{C}^\top.
$$

\paragraph{Decoupling Query Response via Diagonal Approximation.} 

The term $K_\mathcal{C} \Lambda_Q K_\mathcal{C}^\top$ is an $m \times m$ matrix representing the covariance of key responses to future queries. Its off-diagonal element $k_i^\top \Lambda_Q k_j$ measures the correlation between the responses of token $i$ and token $j$. To further decouple the tokens, we approximate this matrix by its diagonal:
$$
    K_\mathcal{C} \Lambda_Q K_\mathcal{C}^\top \approx \operatorname{Diag}(\kappa_i)_{i \in \mathcal{C}}, \quad \text{where } \kappa_i = k_i^\top \Lambda_Q k_i.
$$

The scalar $\kappa_i$ quantifies the variance of the $i$-th token's logit under the future query distribution. A larger $\kappa_i$ indicates that the token's alignment with queries is highly variable, making it potentially more informative than a token with a constant alignment. Ignoring off-diagonal correlations is equivalent to assuming independent key responses under the query prior, a necessary simplification for efficient scoring that is prevalent in similar analyses.

Combining the two diagonal approximations, the inner term becomes a diagonal matrix of scalar weights:
$$
    D_\mathcal{C} (K_\mathcal{C} \Lambda_Q K_\mathcal{C}^\top) D_\mathcal{C} \approx D_\mathcal{C} \operatorname{Diag}(\kappa_i) D_\mathcal{C} = \operatorname{Diag}(\rho_i^2 \kappa_i)_{i \in \mathcal{C}}.
$$
Defining the combined importance weight for token $i$ as 
$$
w_i = \frac{1}{d_k} \rho_i^2 \kappa_i = \frac{1}{d_k} [\alpha_i^\star(1 - \alpha_i^\star)]^2 (k_i^\top \Lambda_Q k_i),
$$
the approximated objective becomes:
\begin{equation}
    \mathcal{L}_\mathcal{C}^{\text{Jac}} \approx \frac{1}{2} \log \det \left( I + \Sigma_{\text{noise}}^{-1} U_\mathcal{C} \operatorname{Diag}(w_i)_{i \in \mathcal{C}} U_\mathcal{C}^\top \right) = \frac{1}{2} \log \det \left( I + \Sigma_{\text{noise}}^{-1} \sum_{i \in \mathcal{C}} w_i u_i u_i^\top \right).
\end{equation}
Assuming isotropic noise $\Sigma_{\text{noise}} = \sigma^2 I$ and absorbing constants into $w_i$, this reduces to the standard form of maximizing the log-determinant of a sum of rank-one matrices.

\paragraph{Leverage Score Selection.} Similar to CapKV~\cite{yang2026rethinkingkvcacheeviction}, instead of combinatorially searching for the optimal subset $\mathcal{C}$, we define a global capacity matrix over the entire candidate pool $\mathcal{H}_t$:
\begin{equation}
    A_{\text{Jac}} = I + \sum_{j \in \mathcal{H}_t} w_j u_j u_j^\top.
\end{equation}
We then select tokens greedily based on their marginal contribution to the log-determinant, which is approximated by the statistical leverage score using the matrix determinant lemma ($ \log \det(A + wu u^\top) - \log \det(A) = \log(1 + w u^\top A^{-1} u) \approx w u^\top A^{-1} u$ for small $w$):
\begin{equation}
    \text{score}_i = w_i \cdot u_i^\top A_{\text{Jac}}^{-1} u_i.
\end{equation}
This final form, used in Algorithm~\ref{alg:jacap}, elegantly combines the theoretically derived nonlinear importance weight $w_i$ with the structural diversity term $u_i^\top A^{-1} u_i$, providing a principled and efficient eviction criterion.

\subsection{Computational Complexity Analysis}
\label{app:jacap_cca}
We analyze the computational complexity of \textsc{Jacap} for a single eviction step with $N$ candidate tokens and head dimension $d_h = d_k = d_o$. The computation consists of three stages.

First, the weight computation requires evaluating the attention weights $\alpha_i^\star$, which costs $O(N d_h)$ for the query-key products and softmax normalization. The query response variance
$$
\kappa_i = k_i^\top \Lambda_Q k_i
$$
requires $O(N d_h^2)$ when $\Lambda_Q$ is full, and reduces to $O(N d_h)$ under a diagonal approximation. Since we adopt a diagonal covariance in practice, the overall cost of computing $w_i$ is $O(N d_h)$.

Second, constructing the Jacobian capacity matrix
$$
A_{\mathrm{Jac}} = \sum_{j=1}^N w_j u_j u_j^\top
$$
requires $O(N d_h^2)$, as each rank-one update costs $O(d_h^2)$. The inversion of the resulting $d_h \times d_h$ matrix further incurs $O(d_h^3)$ complexity. The total cost of this stage is therefore $O(N d_h^2 + d_h^3)$.

Finally, the leverage-score evaluation
$$
u_i^\top A_{\mathrm{Jac}}^{-1} u_i
$$
requires $O(d_h^2)$ per token, leading to a total complexity of $O(N d_h^2)$ for all candidates. Combining all stages, the overall complexity per layer and eviction step is
$$
O(N d_h^2 + d_h^3),
$$
which is dominated by the capacity matrix construction and inversion.

Compared with heuristic approaches such as \textsc{SnapKV}, \textsc{H2O}, \textsc{KeyDiff}, and \textsc{KNorm}, whose complexity is typically $O(N d_h)$ or $O(N)$ due to independent per-token scoring, \textsc{Jacap} introduces additional quadratic dependence on the head dimension through inter-token matrix interactions. However, this complexity is shared by other capacity-based approaches such as \textsc{CapKV}, which also rely on leverage-score computations over a capacity matrix.

In practice, the additional overhead remains moderate because the attention head dimension is usually small (e.g., $d_h = 128$), and the matrix operations are highly parallelizable on modern GPUs. As demonstrated in our runtime experiments, the improved retention quality in high-compression regimes justifies this additional computational cost.
\section{More Experiment Results}
Our all experiments were conducted on $4 \times$ NVIDIA RTX Pro 6000 GPUs with 1024\,GB system memory under Ubuntu 22.04.

\subsection{Runtime Efficiency Analysis}
\label{app:runtime_ana}

\begin{figure}[htbp]
  \centering
  \begin{subfigure}[b]{0.49\linewidth}
    \includegraphics[width=\linewidth]{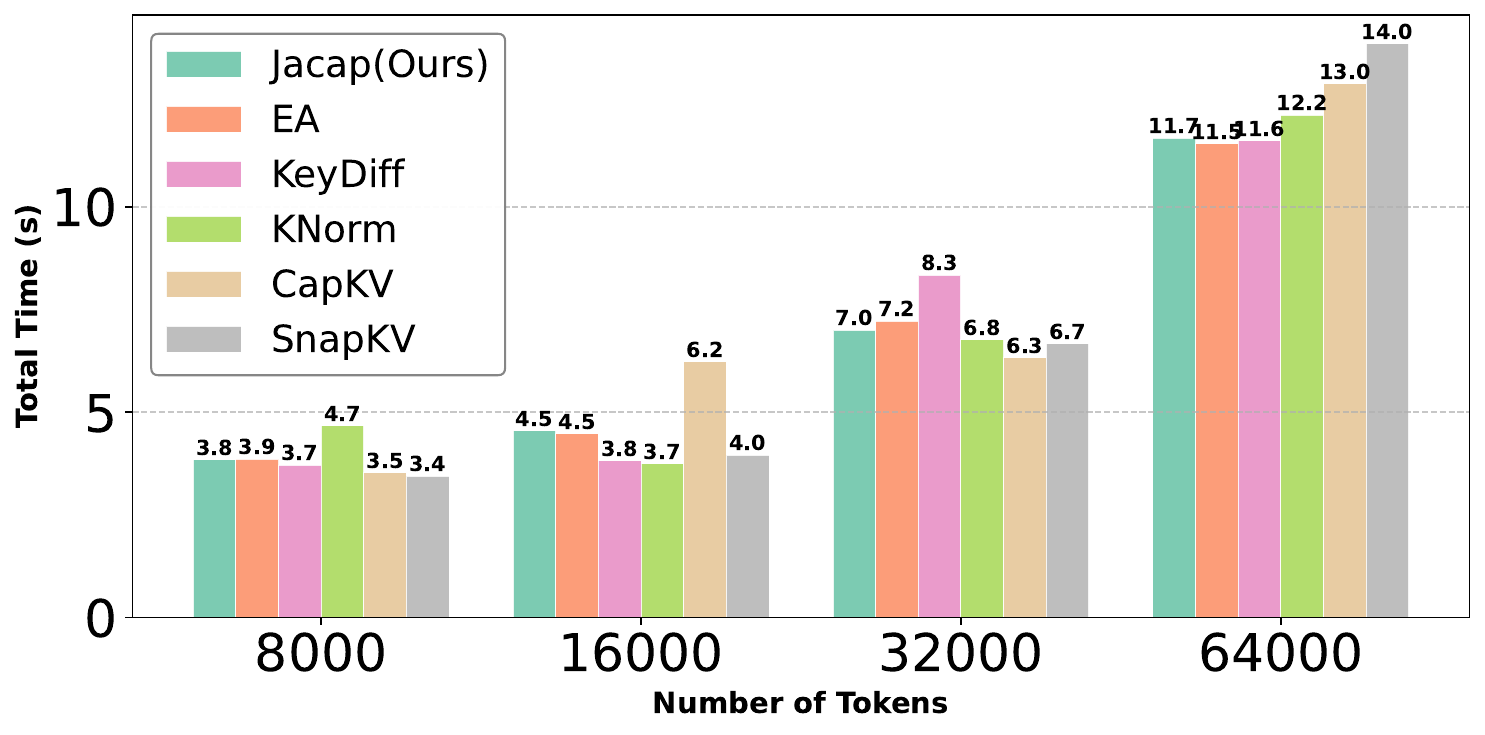}
    \caption{$C.R. = 0.6$}
    \label{fig:sub-a}
  \end{subfigure}
  \hfill
  \begin{subfigure}[b]{0.49\linewidth}
    \includegraphics[width=\linewidth]{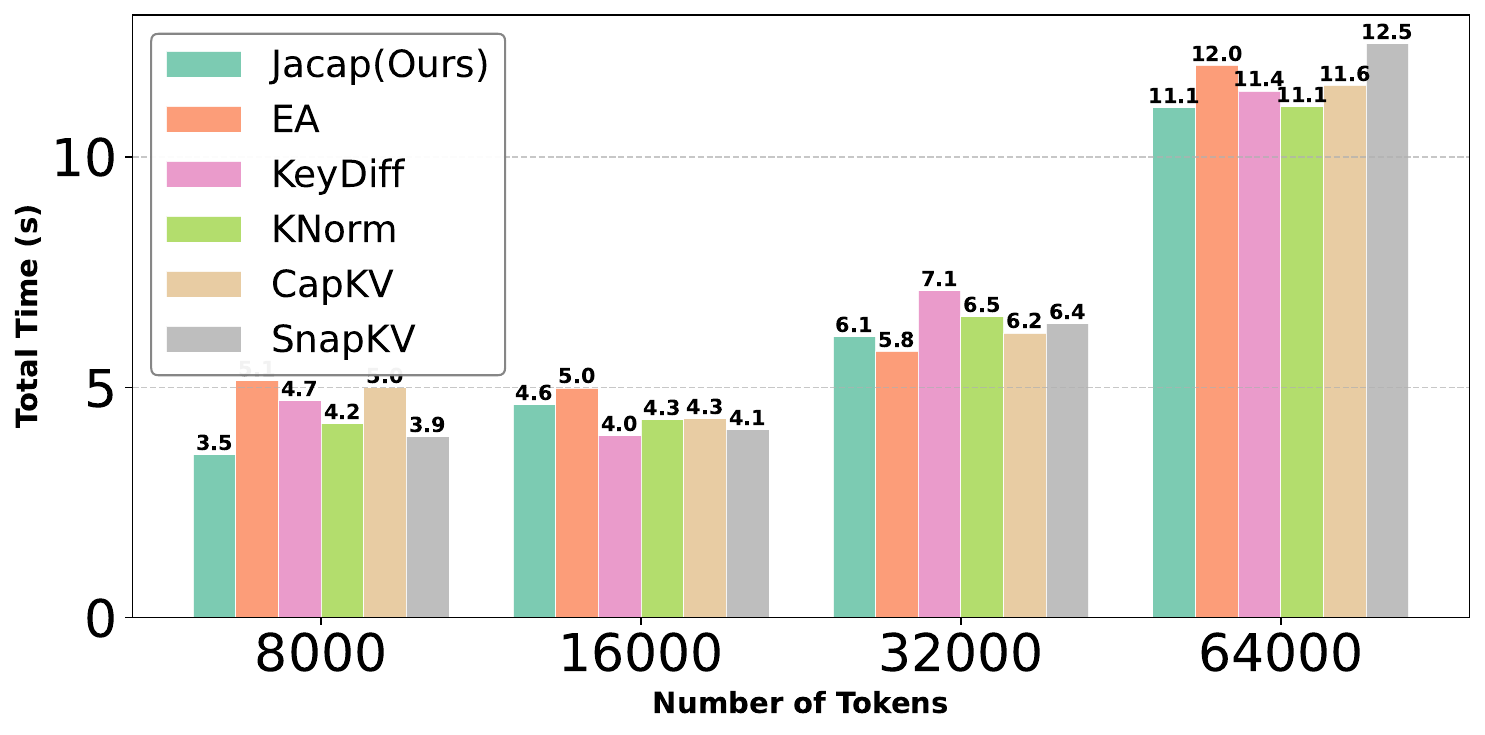}
    \caption{$C.R. = 0.8$}
    \label{fig:sub-b}
  \end{subfigure}
  \caption{Runtime comparison of generating 100 tokens with Qwen3-8B on a single RTX 6000 GPU under varying input context lengths and compression ratios (C.R.)}
  \label{fig:runtime}
\end{figure}

To evaluate the practical overhead of our proposed method, we confirm the runtime efficiency of \textsc{Jacap} against representative baselines. We measured the total generation time for generating 100 tokens using the Qwen3-8B model on a single NVIDIA RTX pro6000 GPU. The evaluation covers increasing input context lengths from 8k to 64k tokens under two compression ratios (C.R. = 0.6 and 0.8).

The results are summarized in Figure~\ref{fig:runtime}. We observe that \textsc{Jacap} exhibits runtime performance that is highly comparable to existing methods across all settings.

The analysis in Appendix~\ref{app:jacap_cca} indicates a theoretical complexity of $O(N d_h^2 + d_h^3)$ for \textsc{Jacap}. These empirical results suggest that in practice, with typical head dimensions (e.g., $d_h=128$), the matrix operations are efficiently handled by the GPU, making the actual overhead negligible compared to the overall inference process. Therefore, \textsc{Jacap} delivers significant performance gains in high-compression regimes without incurring a prohibitive computational cost.

\subsection{More Experiment Result on NIAH}
\label{app:mer_niah}
\begin{figure}[htbp]
  \centering
  \includegraphics[width=\linewidth]{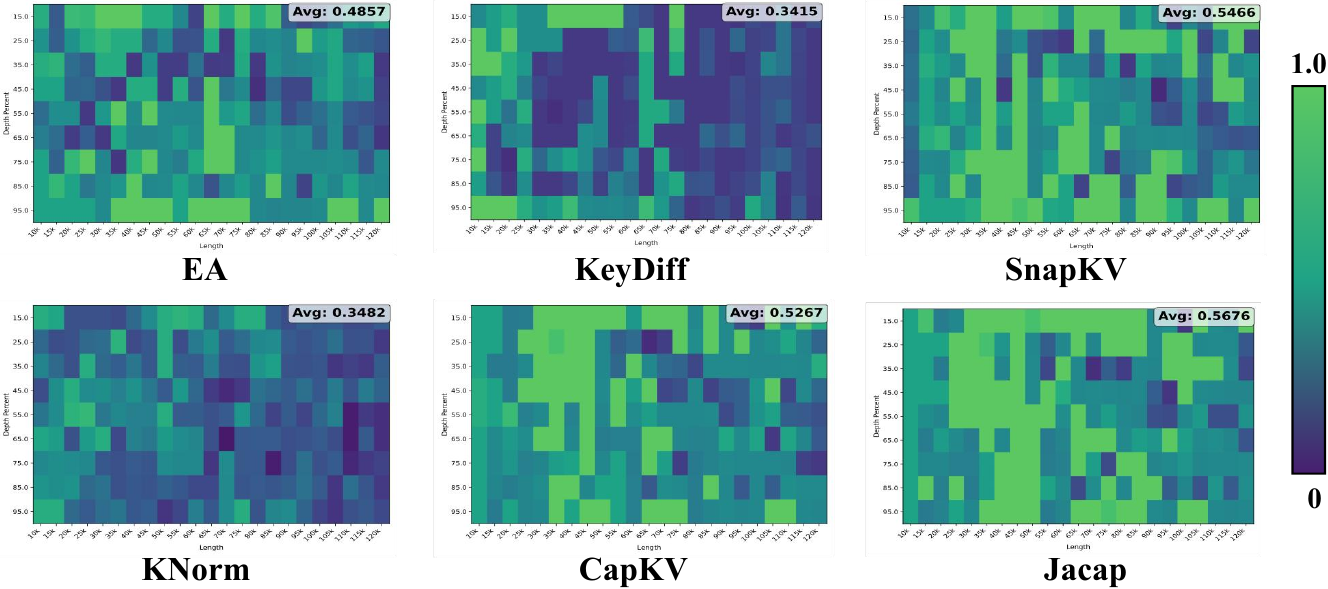}
  \caption{
  Performance of \textsc{Jacap} and baseline KV-cache eviction methods on the NIAH benchmark at compression ratio 0.5 using Qwen3-8B.
  }
\label{fig:niah_50}
\end{figure}

To provide a more comprehensive evaluation across different sparsity regimes, we further evaluate the Needle-in-Haystack (NIAH) performance under a moderate compression ratio of 0.5 using the Qwen3-8B model. The results are shown in Figure~\ref{fig:niah_50}.

Consistent with the findings under higher compression in the main text, \textsc{Jacap} achieves the best overall performance. The heatmaps reveal that while purely structural methods like \textsc{KeyDiff} and \textsc{KNorm} fail significantly at long contexts, and attention-based baselines like \textsc{SnapKV} begin to degrade at extreme lengths (e.g., >100k tokens) and deeper positions, \textsc{Jacap} maintains remarkable robustness. It consistently retrieves the needle across the entire spectrum of context lengths and depths, demonstrating that incorporating nonlinear sensitivity effectively identifies and preserves crucial information even when it is sparsely located deep within very long sequences.

\newpage

\end{document}